\documentclass[sigconf]{acmart}
\AtBeginDocument{%
  \providecommand\BibTeX{{%
    \normalfont B\kern-0.5em{\scshape i\kern-0.25em b}\kern-0.8em\TeX}}}

\copyrightyear{2023}
\acmYear{2023}
\setcopyright{rightsretained}
\acmConference[MADiMa '23]{Proceedings of the 8th International Workshop on Multimedia Assisted Dietary Management}{October 29, 2023}{Ottawa, ON, Canada}
\acmBooktitle{Proceedings of the 8th International Workshop on Multimedia Assisted Dietary Management (MADiMa '23), October 29, 2023, Ottawa, ON, Canada}
\acmDOI{10.1145/3607828.3617795}
\acmISBN{979-8-4007-0284-6/23/10}

\makeatletter
\gdef\@copyrightpermission{
  \begin{minipage}{0.3\columnwidth}
   \href{https://creativecommons.org/licenses/by/4.0/}{\includegraphics[width=0.90\textwidth]{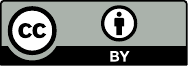}}
  \end{minipage}\hfill
  \begin{minipage}{0.7\columnwidth}
   \href{https://creativecommons.org/licenses/by/4.0/}{This work is licensed under a Creative Commons Attribution International 4.0 License.}
  \end{minipage}
  \vspace{5pt}
}
\makeatother

\usepackage{pifont}% http://ctan.org/pkg/pifont

\newcommand{\xmark}{\ding{55}}%
\newcommand{\cmark}{\ding{51}}%
\begin{document}

\title{An Improved Encoder-Decoder Framework for Food Energy Estimation}

\author{Jack Ma}
\email{ma699@purdue.edu}
\affiliation{%
  \institution{Purdue University}
  \city{West Lafayette}
  \state{Indiana}
  \country{USA}
}

\author{Jiangpeng He}
\email{he416@purdue.edu}
\affiliation{%
  \institution{Purdue University}
  \city{West Lafayette}
  \state{Indiana}
  \country{USA}
}

\author{Fengqing Zhu}
\email{zhu0@purdue.edu}
\affiliation{%
  \institution{Purdue University}
  \city{West Lafayette}
  \state{Indiana}
  \country{USA}
}

\begin{abstract}
  Dietary assessment is essential to maintaining a healthy lifestyle. Automatic image-based dietary assessment is a growing field of research due to the increasing prevalence of image capturing devices (e.g. mobile phones). In this work, we estimate food energy from a single monocular image, a difficult task due to the limited hard-to-extract amount of energy information present in an image. To do so, we employ an improved encoder-decoder framework for energy estimation; the encoder transforms the image into a representation embedded with food energy information in an easier-to-extract format, which the decoder then extracts the energy information from. To implement our method, we compile a high-quality food image dataset verified by registered dietitians containing eating scene images, food-item segmentation masks, and ground truth calorie values. Our method improves upon previous caloric estimation methods by over 10\% and 30 kCal in terms of MAPE and MAE respectively. 
\end{abstract}




\maketitle
\section{Introduction}
Dietary assessment is the process of evaluating an individual's dietary intake, which plays an important role in maintaining a healthy lifestyle, such as identifying nutrient deficiencies and reducing the risk of metabolic disorders. However, traditional methods for dietary assessment~\cite{six2010traditional}, such as self-reporting or filling out questionnaires for dietitians to assess, are often burdensome and time-consuming.
% \begin{figure}[h]
%     \centering
%     \includegraphics[width=\linewidth]{graphs/ex.png}
%     \vspace{-1cm}
%    \caption{Outline of food energy estimation from a single monocular image. The user takes an image of the meal, which is then passed through a energy estimation model to obtain the energy for the meal.}
%    \label{fig:ex}
%\end{figure}

To mitigate these issues, researchers have turned toward automatic image-based dietary assessment methods~\cite{zhu2010multiview, zhu2015multiview, Kitamura2010foodlog, joutou2009recognition, kong2011recognition, he2020depth, he2021end, shao2021_ibdasystem} due to the increasing prevalence of image-capturing devices such as the mobile phone. The initial image-based dietary assessment studies focus on food recognition to predict what food types are consumed, which have been studied under different real-world scenarios such as fine-grained classification~\cite{arslan2021fine, liu2020food, mao2020visual, rodenas2022learning, mao2021_nutri_hierarchy}, long-tailed classification~\cite{he2022long, gao2022dynamic_LTingredient, he2023singlestage} and even for continual lifelong learning~\cite{He_2021_ICCVW, ILIO, raghavan2023online, he2023longtailed, he2022_expfree}. Nonetheless, simply identifying the types of food consumed does not provide any information on caloric or energy intake, which is crucial for a comprehensive dietary assessment. Therefore, the most recent work focuses on food portion size estimation, which aims to predict how much energy is consumed given an eating occasion image as input. These methods use images of the meal to estimate food portions or energy and are faster than traditional dietary assessment approaches. However, most existing methods require the user to take depth maps~\cite{ando2019depth, chen2012depth, kento2022video}, multiple images~\cite{zhu2010multiview, zhu2015multiview}, or videos~\cite{okamoto2016video, kento2022video, sun2010wearable} of the meal, increasing the user's burden in collecting these images. 

In this work, we focus on the simplest form of image-based dietary assessment: food energy estimation from a single monocular image. This is the easiest form of dietary assessment for users since images are quick and easy to take (e.g. using a smartphone camera). However, extracting energy information directly from an image poses multiple challenges, including (i) the presence of significant noise in images, which obscures the relevant energy information and complicates its extraction, (ii) the inherent limitation of two-dimensional imaging in capturing the third dimension, such as the depth of food items, leading to the loss of key volumetric information after camera projection, and (iii) frequent occlusion of food items behind other objects in the frame, further complicating the accurate retrieval of energy information. Due to these complexities, images in their raw form are ill-suited for the direct extraction of energy-related data.

%Instead of attempting to extract energy information directly from the image, 
To address this problem, Fang et al.~\cite{fang2019nutrients} proposed an encoder-decoder framework for food energy estimation. They first use an encoder to transform the image into a grayscale which represents the per-pixel energy density for the image. This grayscale contains energy information in an easier-to-extract format than the energy information contained in the image (\textit{i.e. }the calorie value for the meal is better encoded into this representation). They then use the decoder to extract the energy from the encoded representation.
%. However, Fang et al.'s method possesses several limitations. 
However, the use of grayscale in~\cite{fang2019nutrients} imposes a bottleneck on the fidelity of energy information. Given that each pixel in a grayscale image can only assume integer values ranging from 0 to 255, there's a limitation on the granularity of the energy data that can be represented. This constraint leads to the loss of nuanced energy information, as energy density values are rounded to the nearest integer and confined within this narrow range.

% Because the encoded representation is tailored specifically to enable better extraction of the energy information via the decoder, a desirable property is that given the grayscale, the decoder can extract the exact calorie value from it. However, a meal can have an infinite number of possible calorie values, but the number of possible grayscales is limited to $256\cdot H\cdot W$ (where $H$ and $W$ are the dimensions of the grayscales). Thus, energy information is bottlenecked at the grayscale (i.e. it is impossible to obtain the exact calorie value from the meal given the most optimal decoder). Additionally, because each pixel in the grayscale can only take on integer values from 0 to 255, energy information is lost when the energy density values in the grayscale are rounded to the nearest integer and squeezed into that range. 

We address these limitations by introducing a new encoded representation structure which enables better energy information encoding and identifying a simpler and well-motivated decoder based on our encoding process (\textbf{Contribution 1}). To implement our method, we compile a high-quality dataset verified by registered dietitians. We validate our method on the dataset and show that it achieves at least 10\% MAPE and 30 kCal MAE increase in calorie estimation accuracy than previous methods (\textbf{Contribution 2}).

\section{Related Work}
\label{sec:related work}
In this section, we summarize and review existing methods for automatic image-based energy estimation. We first outline methods that employ depth maps, then highlight multiview/continuous stream methods, and finally summarize single monocular image-based methods in which our work lies.\\\\
\textbf{Depth-View}:
Depth view methods~\cite{ando2019depth, chen2012depth} use a depth map of the meal for energy estimation. The depth map provides the depth of food items in the meal and thus contains much of the extra dimensional information needed that is not captured in a two-dimensional image. Through this extra information, the depth map is usually used to generate a 3D voxel representation~\cite{shao2023endtoend} of the meal from which food volume and energy can then be estimated. However, ordinary consumer-level technology (e.g. phones) is unable to obtain the depth map for the meal, making these approaches often elusive to the average user.\\\\
\textbf{Multiview/Continuous Stream}
Many works use multiview or continuous stream sources of the meal for calorie estimation. Multiview methods~\cite{pouladzadeh2014multiview, zhu2010multiview} usually use many images of the same meal taken at different views and angles. More information about the meal leads to a better representation For example, Pouladzadeh et al.~\cite{pouladzadeh2014multiview} capture top-view and side-view images of the meal. Continuous stream methods~\cite{kento2022video, okamoto2016video, sun2010wearable} use a continuous source of data, usually videos, to perform calorie estimation on. For example, Adachi and Yanai~\cite{kento2022video} use a video of the user eating the meal comprising rgb images and associated depth maps to measure the energy of each food item the user eats. These methods are often burdensome for the user to gather the much bigger and complex data of the meal.\\\\
\textbf{Single Monocular}
Energy estimation from a singular monocular image is the simplest form of energy estimation. Many methods go the simple and intuitive route of "portioning and summing" as done in traditional methods, \textit{i.e. }extracting the portion (volume) and type of the food items from the image and using that information to calculate the calories~\cite{myers2015monocular, okamoto2016monocular}. However, volume extraction is difficult because there is no access to the depth of the food items in the 2-dimensional image.

More recently, researchers have looked towards reformulating the energy estimation problem as an encoder-decoder pipeline~\cite{shao2021cd, fang2018icip, fang2019nutrients}. Because extracting energy information from the image directly is a challenging task, these methods employ an encoder which embeds energy information into the image, transforming it into an encoded representation which the decoder can then extract the energy information from easier~\cite{shao2021cd, fang2018icip, fang2019nutrients}. Fang et al.~\cite{fang2019nutrients} compute encoded representations in the form of grayscales for all the images, then train a Conditional GAN (cGAN)~\cite{isola2017gan} on these image/grayscale pairs to output a generated grayscale given an input image. They then employ a regression network using VGG16~\cite{simonyan2015vgg} to reduce the grayscale to a single value, which represents the calories for the image. Shao et al.~\cite{shao2021cd} input the image along with the encoded representation into the decoder along with utilizing layer normalization (LN)~\cite{ln} and group normalization (GN)~\cite{gn}, improving performance. However, these methods suffer in performance as they are bottlenecked at the grayscale, which simply cannot encode enough energy information in an efficient manner. We further advance the encoder-decoder architecture outlined above and improve upon the encoded representation in this paper.
\section{Dataset}
\label{sec:dataset}
\begin{figure}[h]
    \centering
    \includegraphics[width=\linewidth]{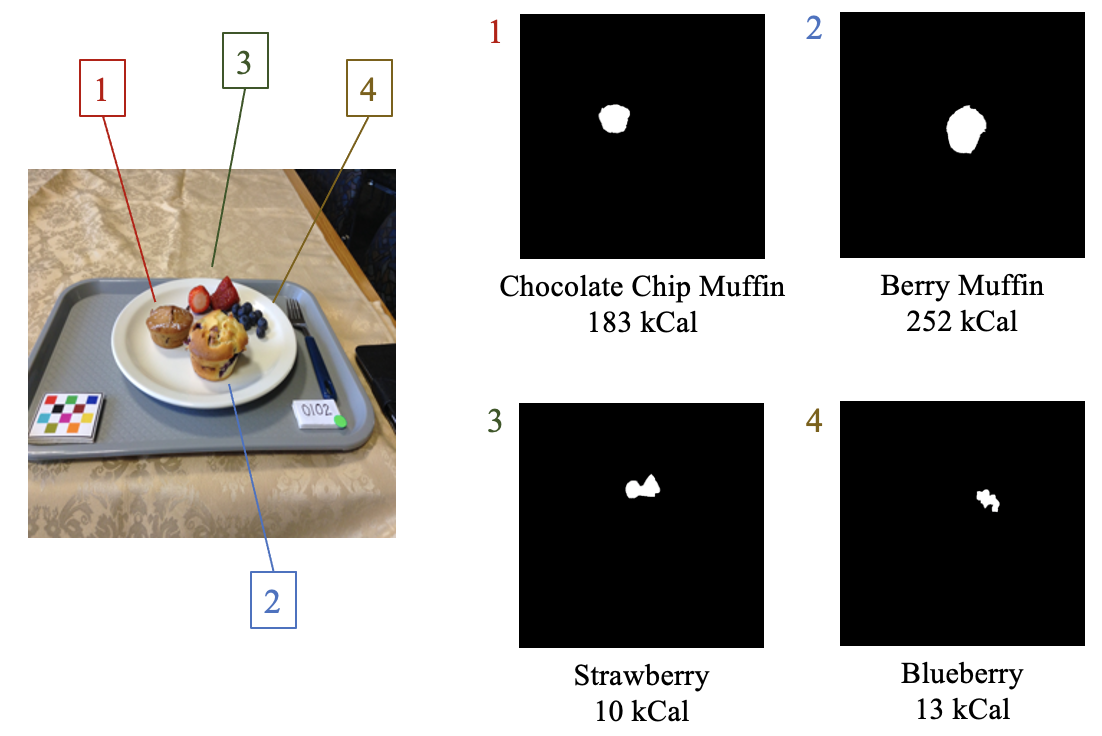}
    \caption{Sample data instance in our dataset. It contains the image taken of the meal, along with the food type, calories, and segmentation mask for each food item in the image.}
    \label{fig:data_instance}
\end{figure}
\begin{figure*}[t]
    \centering
    \includegraphics[width=\linewidth]{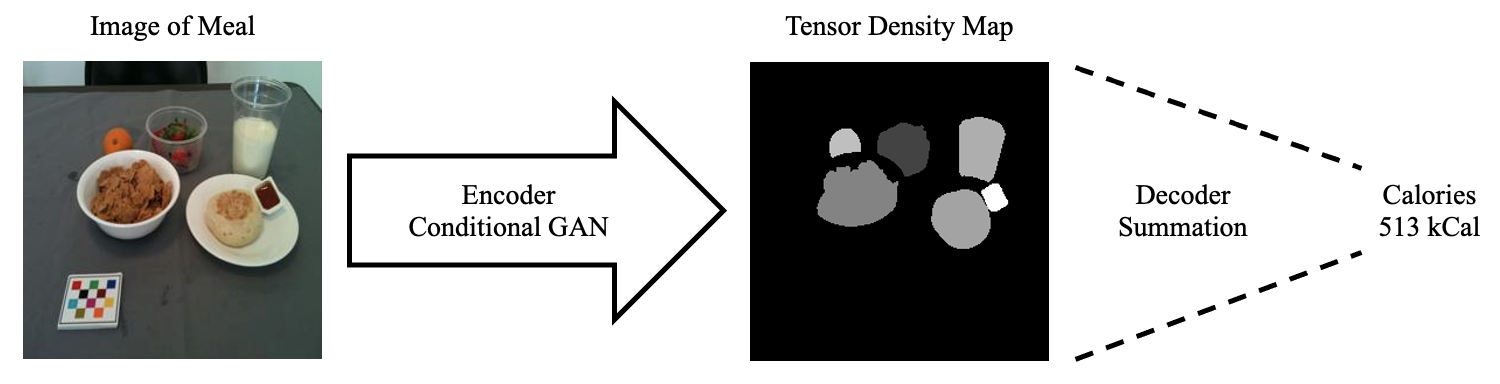}
    \caption{Overview of our model, consisting of (1) a Conditional GAN encoder which takes as input the image and outputs the generated encoded representation (tensor density map, converted the grayscale above for viewing) and (2) our summation decoder which sums up all elements in the density map to obtain the estimated calorie value.}
    \label{fig:model}
\end{figure*}
\begin{figure*}[t]
    \centering
    \includegraphics[width=\linewidth]{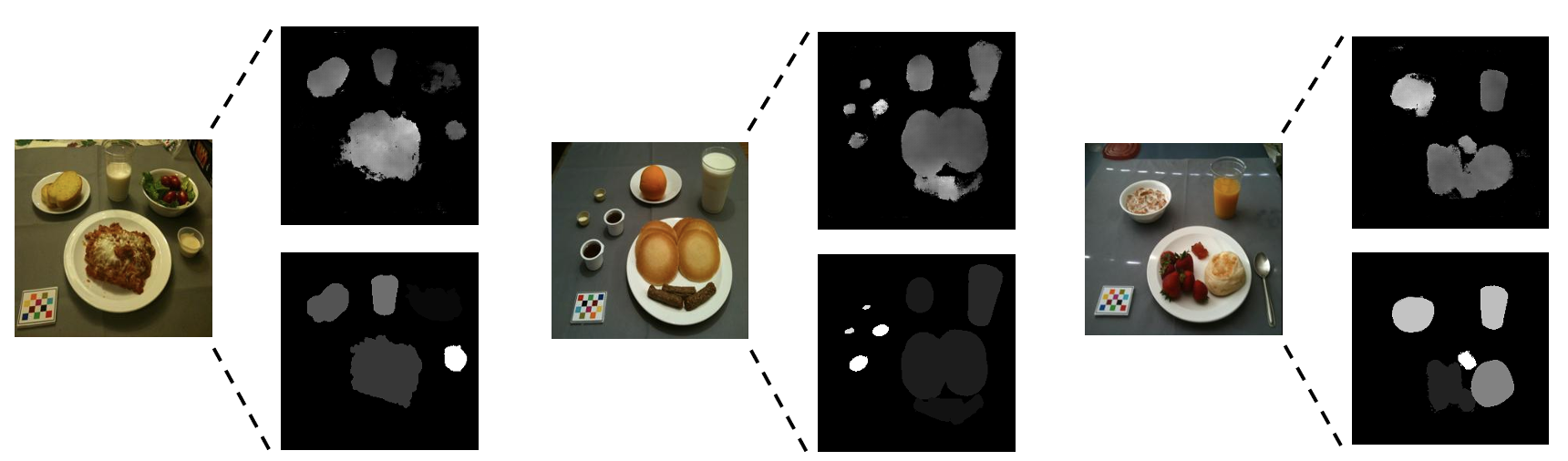}
    \caption{Exemplary tensor density maps generated by the cGAN encoder (top right), along with the ground truth tensor density map (bottom right) and original image (left) for comparison. Note that these density maps are represented as tensors that cannot be viewed pictorially, so they are instead displayed above as grayscales by taking the relative intensities of the values in the tensor.}
    \label{fig:tensor}
\end{figure*}

Existing nutritional datasets include the eating occasion images collected from studies conducted in~\cite{he2020depth, fang2018icip, fang2019nutrients, shao2021cd}. To collect these datasets, each study distributes pre-weighted food items to a selected group of individuals. Each individual then makes their meals for breakfast, lunch, and dinner using the distributed food items and sends a picture of each meal before eating. Food item-wise calories are then calculated using the known weight and type of each food item and are verified by registered dietitians. Each instance in these datasets comprises (i) a picture of the meal (taken while sitting to minimize angle and zooming distortions),  (ii) food type, (iii) calorie value (provided by the registered dietitians), and (iv) corresponding grayscale segmentation mask where each single food item is shown as white pixels. Fig.~\ref{fig:data_instance} shows an exemplary eating occasion image with corresponding annotations in these datasets.

However, these datasets possess several limitations. First, due to the intensive and time-consuming process of distributing food items to individuals and parsing the returned information, each dataset contains a limited number of images (less than 100 in total). Second, food types and calories are stored across several spreadsheets, while the images/segmentation masks are located in a separate dataset folder, and the only connection between them is the date of the meal (represented by a column in the spreadsheets and the file name in the dataset folder). Third, there are some inaccuracies such as missing calorie values for many of the food items due to human errors.

To fix these issues, we first regularize all images to dimensions 256 by 256 in both datasets and combine them into a more extensive dataset. We then manually match the food item-wise calorie value with each segmentation mask. Lastly, we prune the dataset to remove any of the inaccuracies mentioned previously. Our summarized final dataset comprises data from 175 different meals spanning 21 food categories over all meals of the day (breakfast, lunch, dinner). We split the dataset randomly into training, testing, and validation partitions (70\%, 20\%, and 10\% of the dataset respectively) and use these for all our experiments. Note that because all images are taken while sitting right in front of the meal, there is little variance between images in terms of angle and zooming. Thus, we do not apply rectification to the images when implementing our method. 

\section{Method}
Our model employs an encoder-decoder framework for food energy estimation from a single monocular image. Given an input image, we first feed the image into the encoder, which outputs an encoded representation that is embedded with energy information. The decoder is tasked with extracting the energy information from the encoded representation. Energy information is always measured in kCal in this paper. We provide a diagram of our model architecture in Fig.~\ref{fig:model}.

\subsection{Encoder}
It is difficult to extract energy information directly from an image. The encoder serves to transform the image into an encoded representation which contains energy information in a form which we can easily extract. We use a density map as the encoded representation which provides the energy density per-pixel in the image. Before we proceed with details regarding the encoder itself, we will first overview how to generate the density map for an image.
\subsubsection{Density Map Generation}
\label{sec:density-gen}
As outlined in~\ref{sec:dataset}, an image can be divided into its corresponding segmentation masks, where each segmentation mask represents the location of a food item in the image. Given an image, we use the segmentation masks and calorie value associated with each food item to generate the density map. 

Let $n$ denote the number of food items in the image, $\{s_1,...,s_n\}$ denote the segmentation masks, $\{c_1,...,c_n\}$ denote the corresponding calorie values, and $H, W$ denote the height and width of the segmentation masks. For the segmentation mask and calorie value pair $(s_i, c_i)$ associated with a food item, we first generate the food item density map $d_i$ (represented as a tensor also having dimensions $H$ by $W$) by spreading the calories $c_i$ over $s_i$. Formally, let $w_i$ denote the number white pixels in the segmentation mask; then 
\begin{equation}
\label{eq:food_item_grayscale_calc}
    d_i[h,w]=\begin{cases}
	    0 & s_i[h,w]=0\\
        c_i/w_i & \text{otherwise}
	\end{cases}
\end{equation}
for all $1\leq h \leq H$ and $1\leq w \leq W$ (for an image $I$, $I[h,w]$ denotes the pixel value at height $h$ and width $w$).

Once we obtain all food item density maps $d_i$, we simply concatenate all $\{d_i,..,d_n\}$ to obtain the density map $d$. Note that because the segmentation masks are all disjoint, all food item density maps $d_i$ are also disjoint, making concatenation a simple operation. 

Our density map generation possesses several key benefits over~\cite{fang2019nutrients}. First, we represent the density map using a tensor, obviating the need for any rounding/truncating of values in the density map (unlike the grayscale). Second, because elements in tensors are real numbers which can take on an infinite number of values, tensors can hold more energy information than grayscales. In fact, we are able to extract the precise calorie value from the tensor density map, which is impossible for the grayscale. To do so, observe that the sum of all values in the food item density map $d_i$ is $w_i\cdot (c_i/w_i)=c_i$. Since the density map is just the concatenation of all $d_i$, the sum of elements in the density map is precisely $\sum_{i=1}^n c_i$, which is the calorie value of the meal. Thus, the calorie value is simply obtained by summing up all elements in the tensor.

\begin{table}[t]
% \footnotesize
\begin{tabular*}{\columnwidth}{@{\extracolsep{\fill}}l|cc}
\hline
                                       & MAE (kCal)     & MAPE (\%)   \\ \hline
Grayscale~\cite{fang2019nutrients} & 183.5     & 48.5      \\
Image Only~\cite{shao2021cd}      & 287.7     & 61.2   \\
Density Map + Image, LN + GN~\cite{shao2021cd}      & 219.1     & 54.9     \\ 
Density Map + Image, LN~\cite{shao2021cd}      & 208.4     & 58.3     \\ 
\textbf{Ours}                                   & \textbf{150.5} & \textbf{35.7} \\
\hline
\end{tabular*}
\caption{Comparison of methods in terms of mean absolute error (MAE) and mean absolute percent error (MAPE). We include the several implementations Shao et al.~\cite{shao2021cd} provide of their methods, from using only the image as input to the decoder (worse performance) to using both the density map (grayscale) and image as input with layer norm (LN) and layer norm + group norm (LN + GN).}
\label{table:main-results}
\end{table}

\subsubsection{Encoder Model}
In the real world, we are only given the image of the meal and not any segmentation masks, so we need to train an encoder model to learn the mapping between image and density map. Following~\cite{fang2019nutrients}, we use a Conditional Generative Adversarial Network (cGAN)~\cite{isola2017gan}, a conditional generative model that is widely employed in many image to image translation tasks. Once we train the cGAN on image/density map pairs, we can use it to generate the tensor density map given an input image. See Fig.~\ref{fig:tensor} for examples of density maps generated by the cGAN. During the testing phase, we generate the tensor density map for each image using the encoder model and then pass it through the decoder to obtain the estimated calorie value.

\subsection{Decoder}
The decoder is tasked with extracting energy information from the encoded representation generated by the encoder. As mentioned in Sec.~\ref{sec:density-gen}, our density map has the nice property that simply summing up values in the density map yields the calorie value. We thus set our decoder as this simple summation (\textit{i.e. }the decoder sums up all values in the density map). Formally, given the generated tensor density map $d'$, we obtain the calorie value $c$ as
\begin{equation}
\label{eq:food_item_grayscale_calc}
    c=\sum_{i=1}^{H}\sum_{j=1}^{W} d'[i,j]
\end{equation}
where $H$ and $W$ are the dimensions of the image.
We show in Sec.~\ref{sec:exp-abalation} that this decoder performs on par with the more complex regression decoders used in~\cite{fang2019nutrients, fang2018icip, shao2021cd}.

\begin{figure}[t]
    \centering
    \includegraphics[width=\linewidth]{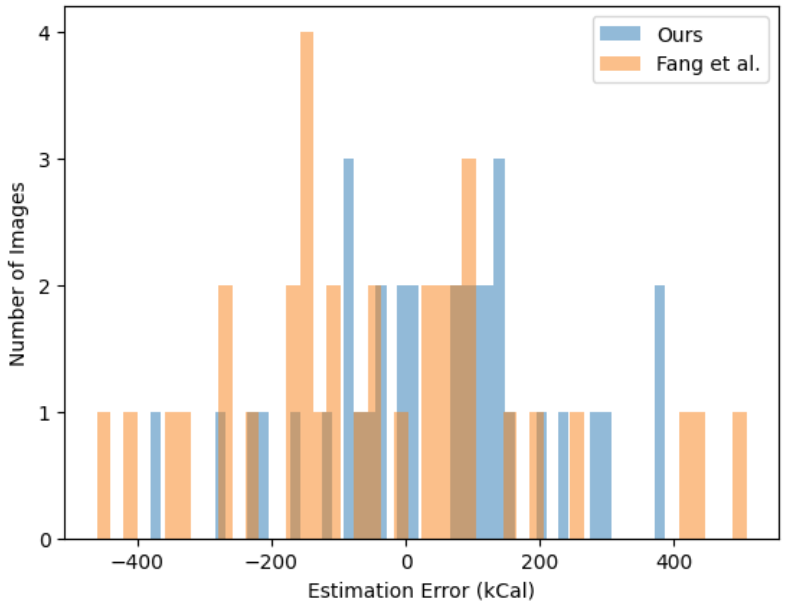}
    \caption{Distribution of the energy estimation errors for our method (blue) in comparison to the next best method (Fang et al.~\cite{fang2019nutrients}, orange). Each data point is calculated by subtracting the ground truth calorie value from the estimated calorie value.}
    \label{fig:hist}
\end{figure}
\section{Experiments}
\label{sec:exp}
\begin{figure*}[h]
    \centering
    \includegraphics[width=\linewidth]{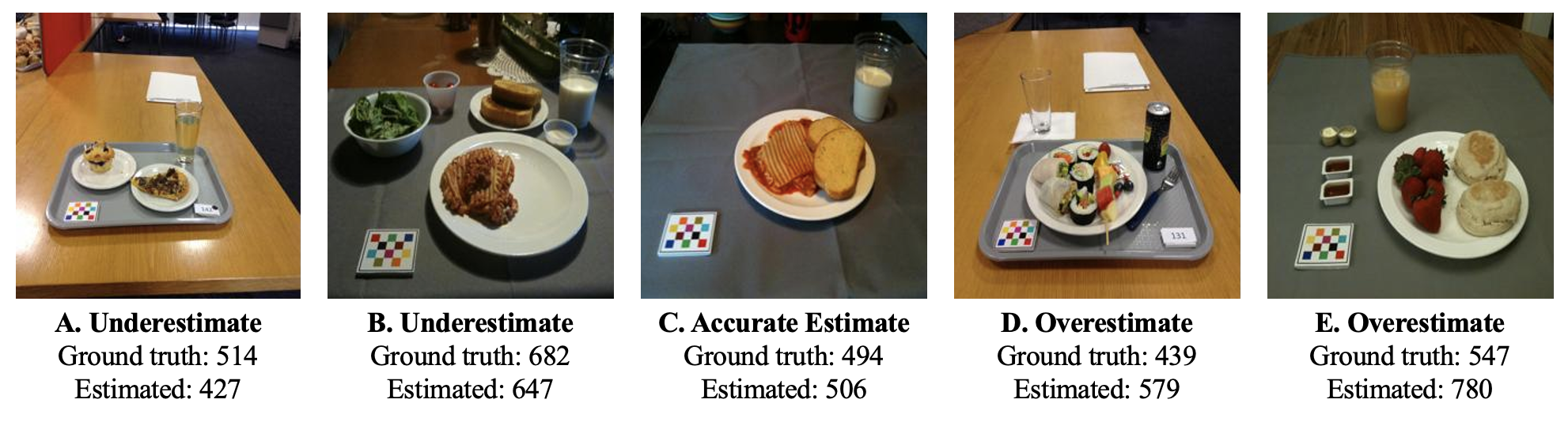}
    \caption{Examples of food energy estimates (kCal) from images in our testing partition.}
    \label{fig:error}
\end{figure*}
In this section, we begin by providing training details for our model in Sec.~\ref{sec:exp-train_detail}, then supply our results in Sec.~\ref{sec:exp-main_res}, and finally perform an abalation study regarding our summation decoder in Sec.~\ref{sec:exp-abalation}. Note that we randomly partition our dataset into training, validation, and testing partitions which we use for all experiments as illustrated in Sec.~\ref{sec:dataset}.

\subsection{Implementation Details}
\label{sec:exp-train_detail}
\subsubsection{Training}
We employ a conditional GAN (cGAN)~\cite{isola2017gan} as our encoder. We train the cGAN for 200 epochs with batch size 1 and the Adam optimizer with initial learning rate 0.0002. Before inputting the image into the cGAN during training, we first randomly flip each image in the training partition along its horizontal or vertical axis. This serves to augment the size of our training dataset by four-fold. Note that this data augmentation process is only applied to the training data—no augmentation is performed during testing. We train the encoder in the previous methods~\cite{fang2019nutrients} with our dataset in the same manner outlined above.

Our simple summation decoder doesn't require any training. As for the neural-network based decoder used in~\cite{fang2019nutrients}, we train a regression model built with either VGG16~\cite{simonyan2015vgg}, Resnet18~\cite{he2018resnet}, or Resnet50~\cite{he2018resnet} for 50 epochs with early validation stopping (\textit{i.e. }if the validation loss does not improve within 20 epochs then we stop the training process).
\subsubsection{Testing}
We test all methods on our dataset testing partition. For each method, we calculate two metrics, mean absolute error (MAE) and mean absolute percent error (MAPE). We input each image $I_i$ in the testing partition into the encoder model to obtain the corresponding encoded representation, which we then pass through the decoder to obtain the estimated calorie value $c_{i}'$. We average $|c_{i}'-c_i|$ and $|c_{i}'-c_i|/c_i\cdot 100$  across all testing data instances to obtain the MAE and MAPE respectively ($c_i$ is the ground truth calorie value for the meal in image $I_i$). The final MAE and MAPE reported in the tables are the average results over five runs of the method.

\subsection{Results}
\label{sec:exp-main_res}
As discussed in Section~\ref{sec:related work}, the majority of existing research relies on additional depth or multi-view images to estimate food energy. This imposes a higher burden on users and serves as a barrier to real-world applications. In this section, we compare our method with the existing approaches using only monocular eating occasion images for food portion estimation including~\cite{fang2019nutrients} and~\cite{shao2021cd}. We report the performance of our method in comparison to the prior works on our introduced dataset in Table~\ref{table:main-results}. As shown in the table, our method achieves a MAE of $150.5$ and MAPE of $35.7\%$, outperforming all previous methods by a large margin. We also provide the distribution of per-data instance estimation errors (estimated calorie value - real calorie value) in Fig.~\ref{fig:hist} in comparison to~\cite{fang2019nutrients} (the next best method in terms of MAE and MAPE). Our method is more accurate and achieves less variance compared to~\cite{fang2019nutrients}. 

However, our method doesn't always obtain better estimation errors than previous methods for every image. Fig.~\ref{fig:error} presents the examples for under estimation, accurate estimation and overestimation on different eating occasion images. This could be due in part to the simple design of the summation decoder. Because we simply sum up the elements in the encoded representation (tensor density map) to obtain the calorie value, if the encoder cannot correctly encode the energy information in this format, the summation decoder will fail to extract the energy information. We examine the usefulness of the summation decoder in the following section.

\begin{table}[t]
% \footnotesize
\begin{tabular*}{\columnwidth}{@{\extracolsep{\fill}}l|ccc}
\hline
                                       & Pretrained & MAE (kCal)     & MAPE (\%)   \\ \hline
VGG16      & \cmark &  166.3 & 38.5   \\
VGG16      & \xmark & 155.5     & 37.9  \\
Resnet18   & \cmark   & 231.8     & 54.7    \\
Resnet18   &  \xmark  & \textbf{149.3} & 35.4   \\
Resnet50   & \cmark    & 173.3     & 37.52    \\
Resnet50   &  \xmark  & 154.0 & \textbf{34.5} \\
\textbf{Ours}                                   & N/A & 150.5 & 35.7 \\
\hline

\end{tabular*}
\caption{Our summation decoder compared against different regression decoders employing VGG16, Resnet18, and Resnet50 (either pretrained on ImageNet or not pretrained).}
\label{table:regression-results}
\end{table}
\subsection{Abalation Study}
\label{sec:exp-abalation}
To investigate whether our summation decoder aids our method or if we would be better off employing the more complex regression decoder in~\cite{fang2019nutrients}, we replace our summation decoder with these regression networks. To remove the noise that comes with training the encoder, we fix a pre-trained encoder on the entire model so that we train the regression models from generated tensors of the encoder. As seen in Table~\ref{table:regression-results}, we see that the summation decoder performs on par with the regression decoders, signaling that the encoder usually encodes the energy information into the encoded representation properly for our summation decoder to extract. Even though there is some improvement in MAE and MAPE when using the regression networks as opposed to the summation decoder, this improvement is marginal (<1\% MAPE and 1.2 kCal MAE) and our decoder is much simpler to use without requiring any regression models. We further observe that at increasing model size (Resnet18, Resnet50, VGG16 from smallest to largest in number of parameters) does not correlate with improving results for the regression-based decoders and pretrained models usually perform subpar compared to their non-pretrained equivalents, which supports our observation that complexity does not necessarily contribute to a more effective decoder.

\begin{table}[t]
% \footnotesize
\begin{tabular*}{\columnwidth}{@{\extracolsep{\fill}}l|cc}
\hline
                                       & MAE (kCal)     & MAPE (\%)   \\ \hline
\textbf{Tensor Density Map (Ours)}       &  \textbf{166.3} & 
\textbf{38.5}   \\
Grayscale       & 183.5    & 48.5  \\
\hline

\end{tabular*}
\caption{Comparison between the tensor density map and grayscale as the encoded representation. Experiments are run using the pipeline from~\cite{fang2019nutrients} because it achieves the best performance out of all methods using a grayscale.}
\label{table:repr-results}
\end{table}

We also explore the effectiveness of using the tensor density map as the intermediate representation as opposed to the grayscale proposed in~\cite{fang2019nutrients}. Sepcifically, we first run~\cite{fang2019nutrients} using their grayscale and then substitute in our tensor density map into their pipeline to observe the portion estimation performance. As shown in Table~\ref{table:repr-results}, using the tensor density map achieves MAE and MAPE of 166.3 kCal and 38.5\% respectively, while using the grayscale instead only achieves MAE and MAPE of 183.5 kCal and 48.5\%. The tensor density map improves both MAE and MAPE results, which aligns with our reasoning in Section~\ref{sec:density-gen} that the tensor density map is able to store more energy information than the grayscale.
\section{Conclusion}
In this paper, we develop an improved encoder-decoder model for calorie estimation given a single monocular image. Our model improves upon previous such models in that (1) our encoded representation is able to contain more energy information before, enabling better encoding of food energy information, and (2) our intuitive decoder is simpler and performs on par with the complex regression decoders in the previous methods. We implement our method on a high-quality eating occasion image dataset containing meal images with associated segmentation masks and calorie values and experimentally show that our method performs better than previous methods by over 10\% and 30 kCal in terms of MAPE and MAE respectively. 

For future work, we believe our encoder-decoder structure has potential to be further improved upon. A promising direction is to explore different forms of the encoded representation to explore a more effective form of encoding food energy than a per-pixel tensor density map. In addition, the lacking of training images pose the major challenge for further improving the portion estimation performance, one possible solution is to use synthetic data as investigated in~\cite{fu2023conditional, pan2022_madima}.

\bibliographystyle{ACM-Reference-Format}
\bibliography{references}

\end{document}